\pdfoutput=1

\documentclass[11pt]{article}

\usepackage{acl}

\usepackage{hyperref}
\usepackage{xcolor}
\usepackage{times}
\usepackage{latexsym}
\usepackage{xcolor,colortbl}
\usepackage{hyperref}
\usepackage{multirow}
\usepackage{tablefootnote}
\usepackage{mathtools}
\usepackage{amsfonts}
\usepackage{amsmath,amssymb}
\usepackage{hhline}
\usepackage{enumitem}
\usepackage{ltablex}
\usepackage{arydshln}
\usepackage{url}

\usepackage[T1]{fontenc}

\usepackage[utf8]{inputenc}

\usepackage{microtype}

\title{Quantified Reproducibility Assessment of NLP Results}

\author{Anya Belz \and Maja Popovi\'c\\
  ADAPT Research Centre \\
  Dublin City University, Ireland \\
  \{anya.belz,maja.popovic\}@adaptcentre.ie \\\And
  Simon Mille \\
  Universitat Pompeu Fabra\\
Barcelona, Spain\\
  simon.mille@upf.edu \\}

\begin{document}
\maketitle
\begin{abstract}
\vspace{-0.15cm}
This paper describes and tests a method for carrying out quantified reproducibility assessment (QRA) that is based on concepts and definitions from metrology. QRA produces a single score estimating the degree of reproducibility of a given system and evaluation measure, on the basis of the scores from, and differences between, different reproductions. We test QRA on 18 system and evaluation measure combinations (involving diverse NLP tasks and types of evaluation), for each of which we have the original results and one to seven reproduction results. The proposed QRA method produces degree-of-reproducibility scores that are comparable across multiple reproductions not only of the same, but of different original studies. We find that the proposed method facilitates insights into causes of variation between reproductions, and allows conclusions to be drawn about what changes to system and/or evaluation design might lead to improved reproducibility.
\end{abstract}

\section{Introduction}

\vspace{-0.125cm}
Reproduction studies are becoming more common in Natural Language Processing (NLP), with the first shared tasks being organised, including REPROLANG \cite{branco-etal-2020-shared} and ReproGen \cite{belz2021reprogen}. In NLP, reproduction studies generally address the following question: \textit{if we create and/or evaluate this system multiple times, will we obtain the same results?}

To answer this question for a given specific system, typically \cite{wieling2018reproducibility,arhiliuc-etal-2020-language,popovic-belz-2021-reproduction} an original study is selected and repeated  more or less closely, before comparing the results obtained in the original study with those obtained in the repeat, and deciding whether the two sets of results are similar enough to support the same conclusions.

This framing, whether the same conclusions can be drawn, involves subjective judgments and different researchers can come to contradictory conclusions: e.g.\ the four papers \cite{arhiliuc-etal-2020-language, bestgen-2020-reproducing,caines-buttery-2020-reprolang,huber-coltekin-2020-reproduction} reproducing \citet{vajjala-rama-2018-experiments} in REPROLANG
all report similarly large differences, but only \citeauthor{arhiliuc-etal-2020-language} conclude that reproduction was unsuccessful. 
There is no standard way of going about a reproduction study in NLP, and different reproduction studies of the same original set of results can differ substantially in terms of their similarity in system and/or evaluation design (as is the case with the  \citet{vajjala-rama-2018-experiments} reproductions, see Section~\ref{sec:results} for details). Other things being equal, a more similar reproduction can be expected to produce more similar results, and such (dis)similarities should be factored into reproduction analysis and conclusions, but NLP lacks a method for doing so. 

Being able to assess reproducibility of results objectively and comparably is important not only to establish that results are valid, but to provide evidence about which methods have better/worse reproducibility and what may need to be changed to improve  reproducibility. To do this, assessment has to be done in a way that is also comparable across reproduction studies of {different} original studies, e.g.\ to develop common expectations of how similar original and reproduction results should be for different types of system, task and evaluation.

In this paper, we (i) describe a method for quantified reproducibility assessment (QRA) directly derived from standard concepts and definitions from metrology which addresses the above issues, and (ii) test it on diverse sets of NLP results. Following a review of related research (Section~\ref{sec:related-research}), we present the method (Section~\ref{sec:framework}), tests and results (Section~\ref{sec:results}), discuss method and results (Section~\ref{sec:discussion}), and finish with some conclusions (Section~\ref{sec:conclusion}).

\section{Related Research}\label{sec:related-research}

The situation memorably caricatured by \citet{pedersen2008empiricism} still happens all the time: you download some code you read about in a paper and liked the sound of, you run it on the data provided, only to find that the results are not the same as reported in the paper, in fact they are likely to be worse \cite{belz2021systematic}. When both data and code are provided, the number of potential causes of such differences is limited, and the NLP field has shared increasingly detailed information about system, dependencies and evaluation to chase down sources of differences. Sharing code and data together with detailed information about them is now expected as standard, and checklists and datasheets have been proposed to standardise information sharing \cite{pineau2020checklist,shimorina2021human}. 

Reproducibility more generally is becoming more of a research focus. There have been several workshops and initiatives on reproducibility, including workshops at ICML 2017 and 2018, the reproducibility challenge at ICLR 2018 and 2019, and at NeurIPS 2019 and 2020, the REPROLANG~\cite{branco-etal-2020-shared} initiative at LREC 2020, and the ReproGen shared task on reproducibility in NLG~\cite{belz2021reprogen}. 

Despite this growing body of research, no consensus has emerged about standards, terminology and definitions. Particularly for the two most frequently used terms, \textit{reproducibility} and \textit{replicability}, multiple divergent definitions are in use, variously conditioned on same vs.\ different teams, methods, artifacts, code, and data.
For example, for \citet{rougier2017sustainable}, \textit{reproducing} a result means running the same code on the same data and obtaining the same result, while \textit{replicating} the result is writing and running new code based on the information provided by the original publication.
For \citet{wieling2018reproducibility}, \textit{reproducibility} is achieving the same results using the same data and methods. 

According to the ACM's definitions \cite{acm2020artifact}, results have been \textit{reproduced} if obtained in a different study by a different team using artifacts supplied in part by the original authors, and \textit{replicated} if obtained in a different study by a different team using artifacts \textit{not} supplied by the original authors. The ACM originally had these definitions the other way around until asked by ISO to bring them in line with the scientific standard (ibid.).

Conversely, in Drummond's view \citeyear{drummond2009replicability}  obtaining the same result by re-running an experiment in the same way as the original is \textit{replicability}, while \textit{reproducibility} is obtaining it in a different way. 

\citet{whitaker2017}, followed by \citet{schloss2018identifying}, defines four concepts rather than two, basing definitions of  \textit{reproducibility, replicability, robustness} and \textit{generalisability} on the different possible combinations of same vs.\ different data and code.

None of these definitions adopt the general scientific concepts and definitions pertaining to reproducibility, codified in the International Vocabulary of Metrology, VIM \cite{jcgm2012international}. One issue is that they all reduce the in principle open-ended number of dimensions of variation between measurements accounted for by VIM to just two or three (code, data and/or team). Another, that unlike VIM, they don't produce comparable results. 

NLP does not currently have a shared approach to deciding reproducibility, and results from reproductions as currently reported are not comparable across studies and can, as mentioned in the introduction, lead to contradictory conclusions about an original study's reproducibility. 
There appears to be no work at all in NLP that aims to estimate \textit{degree} of reproducibility which would allow cross-study comparisons and conclusions.

\section{Metrology-based Reproducibility Assessment}\label{sec:framework}

Metrology is a meta-science: its subject is the standardisation of measurements across all of science to ensure comparability.
Computer science has long borrowed terms, most notably reproducibility, from metrology, albeit not adopting the same definitions (as discussed in Section~\ref{sec:related-research} above). 

In this section, we describe quantified reproducibility assessment (QRA), an approach that is directly derived from the concepts and definitions of metrology, adopting the latter exactly as they are, and yields assessments of the degree of similarity between numerical results and between the studies that produced them. We start below with the concepts and definitions that QRA is based on, followed by an overview of the framework (Section 3.2) and steps in applying it in practice (Section 3.3).

\subsection{VIM Definitions of Repeatability and Reproducibility}\label{sec:vim-def}

The International Vocabulary of Metrology (VIM) \cite{jcgm2012international} defines repeatability and reproducibility as follows (defined terms in bold, see VIM for subsidiary defined terms):

\begin{enumerate}[itemindent=1pt,leftmargin=25pt,topsep=4pt,itemsep=0pt,parsep=3pt,partopsep=0pt]
 \item[2.21] \textbf{measurement repeatability} (or repeatability, for short) is  \textbf{measurement  precision}  under  a  set  of  \textbf{repeatability conditions of measurement}.   

\item[{2.20}] a \textbf{repeatability condition of measurement} (repeatability condition) is a condition  of  \textbf{measurement},  out  of  a  set  of  conditions   that   includes   the   same   \textbf{measurement procedure},   same   operators,   same   \textbf{measuring system},   same   operating   conditions   and   same   location,  and  replicate  measurements  on  the  same  or similar objects over a short period of time. 

\item[{2.25}] \textbf{measurement reproducibility} (reproducibility) is \textbf{measurement   precision}   under   \textbf{reproducibility conditions of measurement}. 

\item[{2.24}] a \textbf{reproducibility condition of measurement} (reproducibility condition) is a condition  of  \textbf{measurement},  out  of  a  set  of  conditions  that  includes  different  locations,  operators,  \textbf{measuring  systems}, etc. A   specification   should   give   the   conditions   changed and unchanged, to the extent practical.

\end{enumerate} 

\noindent In other words, VIM considers repeatability and reproducibility to be properties of measurements (not objects, scores, results or conclusions), and defines them as measurement precision, i.e.\ both are quantified by calculating the precision of a set of measured quantity values. Both concepts are defined relative to a set of conditions of measurement: 
the conditions have to be known and specified for assessment of 
repeatability and reproducibility to be meaningful. In repeatability, conditions are the same, whereas in reproducibility, they differ.

In an NLP context, objects are systems, and measurements involve applying an evaluation method to a system usually via obtaining a sample of its outputs and applying the method to the sample (further details of how concepts map to NLP are provided in Section~\ref{ssec:application}).

\subsection{Assessment framework}\label{ssec:formalisation}

The VIM definitions translate directly to the following definition of repeatability $R^0$ (where all conditions of measurement $C$ are the same across measurements):

\vspace{-.4cm}
\begin{small}
\begin{equation}
\begin{aligned}
\begin{split}
R^0(M_1, M_2, ... M_n) := \,\text{Precision}(v_1, v_2, ... v_n),& \\ \text{where}\, M_i\!: (m, O, t_i, C) \mapsto v_i &
\end{split}
\end{aligned}\label{def:repea}
\end{equation}
\end{small}
\vspace{-.4cm}

\noindent and the $M_i$ are repeat measurements for measurand $m$ performed on object $O$ at different times $t_i$ under (the same) set of conditions $C$, producing measured quantity values $v_i$. Below, the coefficient of variation is used as the precision measure, but other measures are possible. Conditions of measurement are attribute/value pairs each consisting of a name and a value (for examples, see following section). Reproducibility $R$ is defined in the same way as $R^0$ except that condition \textit{values} (but not names) differ for one or more of the conditions of measurement $C_i$:

\vspace{-.4cm}
\begin{small}
\begin{equation}
\begin{aligned}
\begin{split}
R(M_1, M_2, ... M_n) := \,\text{Precision}(v_1, v_2, ... v_n), & \\ \text{where}\,
M_i\!: (m, O, t_i, C_i) \mapsto v_i &
\end{split}
\end{aligned}\label{def:repro}
\end{equation}
\end{small}
\vspace{-.4cm}

\noindent Precision is typically reported in terms of some or all of the following: mean, standard deviation with 95\% confidence intervals, coefficient of variation, and percentage of measured quantity values within $n$ standard deviations. We opt for the coefficient of variation (CV),\footnote{The coefficient of variation (CV), also known as relative standard deviation (RSD) is defined as the standard deviation over the mean, often expressed as a percentage.} because it is a general measure, not in the unit of the measurements (unlike mean and standard deviation), providing a quantification of precision (degree of reproducibility) that is comparable across studies \cite[p.\ 57]{ahmed1995pooling}. This also holds for percentage within $n$ standard deviations but the latter is a less recognised measure, and likely to be the less intuitive for many.

In reproduction studies in NLP/ML, sample sizes tend to be very small (a sample size of 8, one original study plus 7 reproductions, as in Table~\ref{tab:estc-measurements} is currently unique). We therefore need to use de-biased sample estimators: we use the unbiased sample standard deviation, denoted $s^*$, with confidence intervals calculated using a t-distribution, and standard error (of the unbiased sample standard deviation) approximated on the basis of the standard error of the unbiased sample variance $\text{se}(s^2)$ as $\text{se}_{s^2}(s^*) \approx \frac{1}{2\sigma}\text{se}(s^2)$ \cite{rao1973linear}. Assuming measured quantity values are normally distributed, we calculate the standard error of the sample variance in the usual way: $\text{se}(s^2) = \sqrt{\frac{2\sigma^4}{n-1}}$. Finally, we also use a small sample correction (indicated by the star) for the coefficient of variation: $\text{CV}^* = (1+\frac{1}{4n})\text{CV}$ \cite{sokal:rohlf:1969}.\footnote{Code and data are available here: \url{https://github.com/asbelz/coeff-var.}}

Before applying CV$^*$ to values on scales that do not start at 0 (mostly in human evaluations) we shift values to start at 0 to ensure comparability.\footnote{Otherwise CV$^*$ reflects differences solely due to different lower ends of scales.} This means that to calculate the CV$^*$ scores in the tables below, measurements are first shifted.

\subsection{Application of the framework}\label{ssec:application}

Using the defined VIM terms and the notations from Section~\ref{ssec:formalisation}, we can refine the question from the start of this paper as follows: \textit{if we perform multiple measurements of object} O \textit{and measurand} m \textit{under reproducibility conditions of measurement} C$_i$, \textit{what is the precision of the measured quantity values we obtain?}  For NLP, this means calculating the precision of multiple evaluation scores for the same system and evaluation measure.

Focusing here on reproducibility assessment where we start from an existing set of results (rather than a set of experiments specifically designed to test reproducibility), the steps in performing QRA are as follows:

\hspace{-.6cm}\begin{tabular}{p{.475\textwidth}}
\begin{enumerate}[itemsep=0pt,parsep=3pt,topsep=2pt]
    \item For a set of $n$ measurements to be assessed, identify the shared object and measurand.
    \item Identify all conditions of measurement $C_i$ for which information is available for all measurements, and specify values for each condition, including measurement method and procedure.
    \item Gather the $n$ measured quantity values  $v_1, v_2, ... v_n$.
    \item Compute precision for $v_1, v_2, ... v_n$, giving reproducibility score $R$.
    \item Report resulting $R$ score and associated confidence statistics, alongside the $C_i$.
\end{enumerate}
\end{tabular}

\vspace{-.15cm}
\noindent In NLP terms, the object is the ready-to-use system (binaries if available; otherwise code, dependencies, parameter values, how the system was compiled and trained) being evaluated (e.g.\ the NTS-default system variant in Table~\ref{tab:overview}), the measurand is the quantity intended to be measured (e.g.\ BLEU-style modified n-gram precision), and measurement method and procedure capture how to evaluate the system (e.g.\ obtaining system outputs for a specified set of inputs, and applying preprocessing and a given BLEU implementation to the latter). 

VIM holds that reproducibility assessment is only meaningful if the reproducibility conditions of measurement are specified for a given test. Conditions of measurement cover every aspect and detail of how a measurement was performed and how the measured quantity value was obtained. The key objective is to capture all respects in which the measurements to be assessed are \textit{known} to be either the same or different. If QRA is performed for a set of existing results, it is often not possible to  discover every aspect and detail of how a measurement was performed, so a reduced set may have to be used (unlike in experiments designed to test reproducibility where such details can be gathered as part of the experimental design).

The reproducibility and evaluation checklists mentioned in Section~\ref{sec:related-research} \cite{pineau2020checklist,shimorina2021human} capture properties that are in effect conditions of measurement, and in combination with code, data and other resources serve well as a way of specifying conditions of measurement, \textit{if} they have been completed by authors. However, at the present time, completed checklists are not normally available. The following is a simple set of conditions of measurement the information required for which \textit{is} typically available for existing work (we include object and measurand for completeness although strictly they are not conditions, as they must be the same in each measurement in a given QRA test):

\begin{table*}[h!]
    \centering\begin{small}
\setlength\tabcolsep{5pt} 
    \begin{tabular}{|l|c|c|p{3.5cm}|c|c|}
        \hline
       \multirow{2}{*}{System (Object)} & Evaluation measure & N & \multicolumn{1}{c|}{\multirow{2}{*}{Papers reporting results}} & \multicolumn{1}{c|}{\multirow{2}{*}{NLP task}} & \multicolumn{1}{c|}{\multirow{2}{*}{Evaluation type}} \\
        & (Measurand) & scores & & & \\
        \hline
        \hline
        \multirow{4}{*}{PASS} & Clarity & 2 & & & \\
         & Fluency & 2 & \citet{van2017pass}, & \multirow{2}{*}{data-to-text} & \multirow{2}{*}{human, intrinsic} \\
         & Identifiability  & \multirow{2}{*}{2} & \citet{mille:etal:2021}  & & \\
         & of stance &  &  & & \\
        \hline
        mult-base & wf1 & 8 & & &   \\
        mult-word$^-$ & wF1 & 8 & & & \\
        mult-word$^+$ & wF1 & 8 & & & \\ 
        mult-POS$^-$ & wF1 & 8 & \citet{vajjala-rama-2018-experiments}, & &\\
        mult-POS$^+$ & wF1 & 8 & \citet{huber-coltekin-2020-reproduction}, & multilingual essay & metric: intrinsic,\\ 
        mult-dep$^-$ & wF1 & 8 & \citet{arhiliuc-etal-2020-language}, & scoring as text & evaluated against\\
        mult-dep$^+$ & wF1 & 8 & \citet{bestgen-2020-reproducing}, & classification &  single reference\\ 
        mult-dom$^-$ & wF1 & 8 & \citet{caines-buttery-2020-reprolang} & & \\
        mult-dom$^+$ & wF1 & 8 & & & \\ 
        mult-emb$^-$ & wF1 & 8 & & & \\
        mult-emb$^+$ & wF1 & 8 & & & \\ 
        \hline
        \multirow{2}{*}{NTS\_default} & BLEU & 7 & \citet{nisioi-etal-2017-exploring}, & & {metric: intrinsic,}\\
         & SARI & 5 &  Cooper \& Shardlow \citeyearpar{cooper-shardlow-2020-combinmt}, & \multirow{2}{*}{text simplification} & eval.\ against input \\
        \multirow{2}{*}{NTS-w2v\_default} & BLEU & 6 & additional reproduction  & & and/or multiple \\
         & SARI & 4 & study for this paper & & references \\
        \hline
    \end{tabular}
    \caption{Summary overview of the 18 object/measurand combinations taht were QRA-tested for this paper.}\label{tab:overview}
    \end{small}
\end{table*}

\begin{enumerate}[itemsep=0pt,parsep=3pt,topsep=3pt]
    \item \textbf{Object}: the system (variant) being evaluated.\footnote{VIM doesn't define `object' but refers to it as that which is being measured.} \textit{E.g.\ a given MT system.}
    \item \textbf{Measurand}: the quantity intended to be evaluated.\footnote{For definition of {`measurand'} see VIM 2.3.} \textit{E.g.\ BLEU-style n-gram precision or human-assessed Fluency.}
    \item Object conditions:
    \begin{enumerate}[itemsep=0pt,parsep=3pt]
        \item \textbf{System code}: source code including any parameters. \textit{E.g.\ the complete code implementing an MT system.}
        \item \textbf{Compile/training information}: steps from code plus parameters to fully compiled and trained system, including dependencies and environment. \textit{{E.g.\ complete information about how the MT system code was compiled and the system trained.}}
    \end{enumerate}
    \item Measurement method conditions:\footnote{For definition of `measurement method', see VIM 2.5.}
    \begin{enumerate}[itemsep=0pt,parsep=3pt]
        \item \textbf{Method specification}: full description of method used for obtaining values quantifying the measurand. \textit{E.g.\ a formal definition of BLEU.}
        \item \textbf{Implementation}: the method implemented in a form that can be applied to the object in order to obtain measured quantity values. \textit{E.g.\ a full implementation of BLEU.}
    \end{enumerate}
    \item Measurement procedure conditions:\footnote{For definition of `measurement procedure', see VIM 2.6.}
    \begin{enumerate}[itemsep=0pt,parsep=3pt]
        \item \textbf{Procedure}: specification of how system outputs (or other system characteristics) are obtained and the measurement method is applied to them. \textit{E.g.\ running a BLEU tool on system outputs and reference outputs.} 
        \item \textbf{Test set}: the data used in obtaining and evaluating system outputs (or other system characteristics). \textit{E.g.\ a test set of source-language texts and {reference translations.}}
        \item \textbf{Performed by}: who performed the measurement procedure and any additional information about how they did it. \textit{E.g.\ the team applying the BLEU tool, and the run-time environment they used.}
    \end{enumerate}
\end{enumerate}

\noindent The \textit{names} of the conditions of measurement used in this paper are boldfaced above. The \textit{values} for each condition characterise how measurements differ in respect of the condition. In reporting results from QRA tests in the following section, we use paper identifiers as shorthand for each distinct condition value (full details in each case being available from the referenced papers).

\begin{table*}[h!]
   \centering
\setlength\tabcolsep{4.4pt} 
\renewcommand{\arraystretch}{1.25} 
\sffamily\selectfont
    \begin{footnotesize}
    \begin{tabular}{|l|l|c|c|c|c|c|c|c|}
    \hline
        & & \multicolumn{2}{c|}{Measured quantity value} & Sample &&&&\\
        \cline{3-4}
        Object & Measurand & \citeauthor{van2017pass} & \citet{mille:etal:2021} & size & mean & stdev & stdev 95\% CI & CV$^*$ $\downarrow$  \\
        &&(2017)&&&&&&\\
    \hline
    \hline
        \multirow{3}{*}{PASS} & Clarity & 5.64 & 6.30 & 2 & 4.969 & 0.583 & [-2.75, 3.92] & 13.193\\
         & Fluency & 5.36 & 6.14 & 2 & 4.75 & 0.691 & [-3.26, 4.65] & 16.372\\
         & Stance id.\ & 91\% & 97\% & 2 & 93.88 & 5.096 & [-24.05, 34.24] & 6.107\\
    \hline
    \end{tabular}
    \end{footnotesize} 
    \caption{Precision (CV$^*$) and component measures (mean, standard deviation, standard deviation, confidence intervals) for measured quantity values obtained in two measurements for each of the three human-assessed evaluation measures for the \textbf{PASS system}. Columns~6--9 calculated on {shifted} scores (see Section~\ref{ssec:formalisation}).} 
    \label{tab:pass-measurements}
\end{table*}

\begin{table*}
    \centering
\setlength\tabcolsep{4.6pt} 
\renewcommand{\arraystretch}{1.35} 
\sffamily\selectfont
\begin{scriptsize}
\begin{tabular}{|c|c|c|c|c|c|c|c|c|c|c|c|c|c|}
    \hline
     & & \multicolumn{2}{c|}{\multirow{2}{*}{\textit{Object conditions}}} & \multicolumn{2}{c|}{{\textit{Measurement method}}} & \multicolumn{3}{c|}{{\textit{Measurement procedure}}} & \multicolumn{1}{c|}{{Measured}} & \multicolumn{1}{c|}{}\\
    Object & Measurand & \multicolumn{2}{c|}{} & \multicolumn{2}{c|}{{\textit{conditions}}} & \multicolumn{3}{c|}{{\textit{conditions}}} & \multicolumn{1}{c|}{{quantity}} & \multicolumn{1}{c|}{\textit{CV$^*$}} \\
    \cline{3-9}
     &  & {\textit{Code by}} & {\textit{Comp./trained by}} & \textit{Method} & \textit{Implem.\ by} & \textit{Procedure} & \textit{Test set} & \multicolumn{1}{c|}{\textit{Performed by}} &  {value} & \multicolumn{1}{c|}{}\\
    \hline
    \hline
        \multirow{6}{*}{PASS} & \multirow{2}{*}{Clarity} & vdL\&al & vdL\&al & vdL\&al & vdL\&al & vdL\&al & \multicolumn{1}{c|}{vdL\&al} & vdL\&al & 5.64 &\multicolumn{1}{c|}{\multirow{2}{*}{\textit{13.193}}}\\
        \cdashline{3-10}
         &  & vdL\&al & vdL\&al & vdL\&al & M\&al & M\&al & \multicolumn{1}{c|}{vdL\&al} & M\&al & 6.30 &\multicolumn{1}{c|}{}\\
         \cline{2-12}
         & \multirow{2}{*}{Fluency} & vdL\&al & vdL\&al & vdL\&al & vdL\&al & vdL\&al & \multicolumn{1}{c|}{vdL\&al} & vdL\&al & 5.36 &\multicolumn{1}{c|}{\multirow{2}{*}{\textit{16.372}}}\\
        \cdashline{3-10}
         &  & vdL\&al & vdL\&al & vdL\&al & M\&al & M\&al & \multicolumn{1}{c|}{vdL\&al} & M\&al & 6.14 &\multicolumn{1}{c|}{}\\
         \cline{2-12}
         & \multirow{2}{*}{Stance id.} & vdL\&al & vdL\&al & vdL\&al & vdL\&al & vdL\&al & \multicolumn{1}{c|}{vdL\&al} &  vdL\&al & 91\%&\multicolumn{1}{c|}{\multirow{2}{*}{\textit{6.107}}}\\
        \cdashline{3-10}
         &  & vdL\&al & vdL\&al & vdL\&al & M\&al & M\&al & \multicolumn{1}{c|}{vdL\&al} & M\&al  & 96.75\%&\multicolumn{1}{c|}{}\\
    \hline
\end{tabular}
\end{scriptsize}
    \caption{Conditions of measurement for two measurements each for three evaluation measures (measurands) and the \textbf{PASS system}. vdL\&al = \citet{van2017pass}; M\&al = \citet{mille:etal:2021}.}
    \label{tab:pass-conditions} 
\end{table*}

\section{QRA Tests}\label{sec:overview}\label{sec:results}

Table~\ref{tab:overview} provides an overview of the 18 object/ measurand pairs (corresponding to 116 individual measurements) for which we performed QRA tests in this study. For each object/measurand pair, the columns show, from left to right, information about the system evaluated (object), the evaluation measure applied (measurand), the number of scores (measured quantity values) obtained, the papers in which systems and scores were first reported, and the NLP task and type of evaluation involved.

There are three sets of related systems: (i) the (single) PASS football report generator \cite{van2017pass}, (ii) \citet{vajjala-rama-2018-experiments}'s 11 multilingual essay scoring system variants, and (iii) two variants of \citet{nisioi-etal-2017-exploring}'s neural text simplifier (NTS). PASS is evaluated with three evaluation measures (human-assessed Clarity, Fluency and Stance Identifiability), the essay scoring systems with one (weighted F1), and the NTS systems with two (BLEU and SARI). For PASS we have one reproduction study, for the essay scorers seven, and for the NTS systems, from three to six. The PASS reproduction was carried out as part of ReproGen \cite{belz2021reprogen}, the reproductions of the essay-scoring systems and of one of the NTS systems as part of REPROLANG \cite{branco-etal-2020-shared}, and we carried out an additional reproduction study of the NTS systems for this paper.\footnote{Authors of original studies gave permission for their work to be reproduced \cite{branco-etal-2020-shared,belz2021reprogen}.}

The PASS text generation system is rule-based, the essay classifiers are `theory-guided and data-driven' hybrids, and the text simplifiers are end-to-end neural systems. This gives us a good breadth of NLP tasks, system types, and evaluation types and measures to test QRA on.

\subsection{QRA for NTS systems}

The neural text simplification systems reported by \citet{nisioi-etal-2017-exploring} were evaluated with BLEU (n-gram similarity between outputs and multiple reference texts) and SARI (based on word added/retained/deleted in outputs compared to both inputs and reference texts, summing over addition and retention F-scores and deletion Precisions). 

Table~\ref{tab:nts-measurements} shows BLEU and SARI scores for the two system variants from the original paper and the two reproduction studies, alongside the four corresponding CV$^*$ values. In their reproduction, \citet{cooper-shardlow-2020-combinmt} regenerated test outputs for NTS-w2v\_def, but not for NTS\_def, which explains the missing scores in Column~4. The different numbers of scores in different rows in Columns 6--9 are due to our own reproduction using \citeauthor{nisioi-etal-2017-exploring}'s SARI script, but two different BLEU scripts: (i) \citeauthor{nisioi-etal-2017-exploring}'s script albeit with the tokeniser replaced by our own because the former did not work due to changes in the NLTK library; and (ii) SacreBLEU \cite{xu2016optimizing}.

Table~\ref{tab:nts-conditions} shows the conditions of measurement for each of the 22 individual measurements. The measured quantity values for those measurements where \textit{Comp./trained by=Nisioi et al.}\ are identical for the SARI metric (scores highlighted by green/lighter shading and italics), but differ by up to 1.4 points for BLEU (scores highlighted by blue/darker shading). Because \textit{Test set=Nisioi et al.} in all cases, the differences in these BLEU scores can only be caused by differences in BLEU scripts and how they were run. The corresponding CV$^*$ is as big as 0.838 for (just) the four NTS\_def BLEU scores, and 1.314 for (just) the three NTS-w2v\_def BLEU scores, reflecting known problems with non-standardised BLEU scripts \cite{post2018call}. 

\begin{table*}[h!]
   \centering
\setlength\tabcolsep{2.75pt} 
\renewcommand{\arraystretch}{1.25} 
\sffamily\selectfont
    \begin{scriptsize}
    \begin{tabular}{|l|c|c|c|c|c|c|c|c|c|c|c|c|c|c|}
    \hline
        & & \multicolumn{7}{c|}{Measured quantity value} &  &&&&\\
        \cline{3-9}
        Object & Measurand & Nisioi et al.\ & \multicolumn{2}{c|}{Cooper \& Shardlow} & \multicolumn{4}{c|}{this paper} & Sample & mean & stdev & stdev 95\% CI & CV$^*$ $\downarrow$  \\
        \cline{3-9}
        && outputs 1 & outputs 1 & outputs 2 & \multicolumn{2}{c|}{outputs 1} & \multicolumn{2}{c|}{outputs 3} &size &&&&\\
        \cline{3-9}
        && s1 / b1 & s1 / b2 & s1 / b2 & s1 / b3 & s1 / b4 & s1 / b3 & s1 / b4 &&&&&\\
    \hline
    \hline
        \multirow{2}{*}{NTS\_def} & BLEU & 84.51 & 84.50 & 87.46 & 85.60 & 84.20  & 86.61  & 86.20 & 7 & 85.58 & 1.29 & [0.45, 2.13] & 1.562 \\
        \cline{2-14}
         & SARI & 30.65 & 30.65 & 29.13  & \multicolumn{2}{c|}{30.65} &  \multicolumn{2}{c|}{29.96} & 5 & 30.21 & 0.72 & [0.095, 1.34] & 2.487 \\
         \hline
        \multirow{2}{*}{NTS-w2v\_def} & BLEU & 87.50 & -- & 80.75 &  89.36 & 88.10 &89.64  & 88.80 & 6 & 87.36 & 3.502 & [0.92, 6.08] & 4.176 \\
        \cline{2-14}
         & SARI & 31.11 & -- & 30.28 & \multicolumn{2}{c|}{31.11}  & \multicolumn{2}{c|}{29.12} & 4 & 30.41 & 1.02 & [-0.11, 2.15] &  3.572 \\
    \hline
    \end{tabular}
    \end{scriptsize} 
    \caption{Precision (CV$^*$) and component measures (mean, standard deviation, standard deviation confidence intervals) for measured quantity values obtained in multiple measurements of the two \textbf{NTS systems}. Outputs 1 = test set outputs as generated by \citet{nisioi-etal-2017-exploring}; outputs 2 = test set outputs regenerated by \citet{cooper-shardlow-2020-combinmt}; outputs 3 = test set outputs regenerated by the present authors. s1 = SARI script (always the same); b1 = Nisioi et al.'s BLEU script, run by Nisioi et al.; b2 = Nisioi et al.'s BLEU script, run by Cooper \& Shardlow; b3 = Nisioi et al.'s BLEU  script with different version of NLTK tokeniser (see in text), run by the present authors; b4 = SacreBLEU \cite{xu2016optimizing}, run by the present authors.}
    \label{tab:nts-measurements}
\end{table*}

\definecolor{LightAzul}{rgb}{0.9,0.94,1}
\newcolumntype{a}{>{\columncolor{LightAzul}}c}
\definecolor{lightgreen}{rgb}{0.75,1,0.82}
\newcolumntype{g}{>{\columncolor{lightgreen}}c} 
\definecolor{lightred}{rgb}{1,0.76,0.72}
\newcolumntype{r}{>{\columncolor{lightred}}c} 

\begin{table*}[h!]
    \centering
\setlength\tabcolsep{2.25pt} 
\renewcommand{\arraystretch}{1.25} 
\sffamily\selectfont
\begin{scriptsize}
\begin{tabular}{|c|c|c|c|c|c|c|c|c|c|c|c|c|}
    \hline
     & & \multicolumn{2}{c|}{\multirow{2}{*}{\textit{Object conditions}}} & \multicolumn{2}{c|}{{\textit{Measurement method}}} & \multicolumn{3}{c|}{{\textit{Measurement procedure}}} & \multicolumn{1}{c|}{{Measured}} & \multicolumn{1}{c|}{}\\
    Object & Measurand & \multicolumn{2}{c|}{} & \multicolumn{2}{c|}{{\textit{conditions}}} & \multicolumn{3}{c|}{{\textit{conditions}}} & \multicolumn{1}{c|}{{quantity}} & \multicolumn{1}{c|}{\textit{CV$^*$}} \\
    \cline{3-9}
     &  & {\textit{Code by}} & {\textit{Comp./trained by}} & \textit{Method} & \textit{Implem.\ by } & \textit{Procedure} & \multicolumn{1}{c|}{\textit{Test set}} & \multicolumn{1}{c|}{\textit{Performed by}}  & {value} & \multicolumn{1}{c|}{}\\
    \hline
    \hline
        \multirow{12}{*}{NTS\_def} & \multirow{7}{*}{BLEU} & Nisioi et al.\ & Nisioi et al.\ & bleu(o,t) & Nisioi et al.\ & OTE & Nisioi et al.\ &Nisioi et al.\ & \multicolumn{1}{a|}{\textbf{84.51}} &\multicolumn{1}{c|}{\multirow{7}{*}{\textit{ 1.562}}}\\
        \cdashline{3-10}
         &  & Nisioi et al.\ & Nisioi et al.\ & bleu(o,t) & Nisioi et al.\ & OTE & Nisioi et al.\ & Coop.\ \& Shard.\  &\multicolumn{1}{a|}{84.50} &\multicolumn{1}{c|}{}\\
        \cdashline{3-10}
         &  & Nisioi et al.\ & Nisioi et al.\ & bleu(o,t) & $\approx$Nisioi et al.\ & OTE & Nisioi et al.\ & this paper &\multicolumn{1}{a|}{85.60} &\multicolumn{1}{c|}{}\\
        \cdashline{3-10}
         &  & Nisioi et al.\ & Nisioi et al.\ & bleu(o,t) & SacreBLEU & OTE & Nisioi et al.\ & this paper &\multicolumn{1}{a|}{84.20} &\multicolumn{1}{c|}{}\\
        \cdashline{3-10}
         &  & Nisioi et al.\ & Coop.\ \& Shard.\  & bleu(o,t) & Nisioi et al.\ & OTE & Nisioi et al.\ & Coop.\ \& Shard.\  &\textbf{87.46}&\multicolumn{1}{c|}{}\\
        \cdashline{3-10}
         &  & Nisioi et al.\ & this paper & bleu(o,t) & $\approx$Nisioi et al.\ & OTE & Nisioi et al.\ & this paper &\textbf{86.61} &\multicolumn{1}{c|}{}\\
        \cdashline{3-10}
         &  & Nisioi et al.\ & this paper & bleu(o,t) & SacreBLEU & OTE & Nisioi et al.\ & this paper &86.20 &\multicolumn{1}{c|}{}\\
         \cline{2-12}
         & \multirow{5}{*}{SARI} & Nisioi et al.\ & Nisioi et al.\ & sari(o,s,t) & Nisioi et al.\ & OITE & Nisioi et al.\ &Nisioi et al.\ & \multicolumn{1}{g|}{\textbf{\textit{30.65}}} &\multicolumn{1}{c|}{\multirow{5}{*}{\textit{2.487}}}\\
        \cdashline{3-10}
         &  & Nisioi et al.\ & Nisioi et al.\ & sari(o,s,t) & Nisioi et al.\ & OITE & Nisioi et al.\ & Coop.\ \& Shard.\  & \multicolumn{1}{g|}{\textit{30.65}} &\multicolumn{1}{c|}{}\\
        \cdashline{3-10}
         &  & Nisioi et al.\ & Nisioi et al.\ & sari(o,s,t) & Nisioi et al.\ & OITE & Nisioi et al.\ & this paper & \multicolumn{1}{g|}{\textit{30.65}} &\multicolumn{1}{c|}{}\\
        \cdashline{3-10}
         &  & Nisioi et al.\ & Coop.\ \& Shard.\  & sari(o,s,t) & Nisioi et al.\ & OITE & Nisioi et al.\ & Coop.\ \& Shard.\  &\textbf{29.13}  &\multicolumn{1}{c|}{}\\
        \cdashline{3-10}
         &  & Nisioi et al.\ & this paper & sari(o,s,t) & Nisioi et al.\ & OITE & Nisioi et al.\ & this paper &\textbf{29.96} &\multicolumn{1}{c|}{}\\
    \hline
        \multirow{10}{*}{NTS-w2v\_def} & \multirow{6}{*}{BLEU} & Nisioi et al.\ & Nisioi et al.\ & bleu(o,t) &  Nisioi et al.\ & OTE & Nisioi et al.\ &Nisioi et al.\ & \multicolumn{1}{a|}{\textbf{87.50}} &\multicolumn{1}{c|}{\multirow{6}{*}{\textit{4.176}}}\\
        \cdashline{3-10}
         &  & Nisioi et al.\ & Nisioi et al.\ & bleu(o,t) & $\approx$Nisioi et al.\ & OTE & Nisioi et al.\ & this paper & \multicolumn{1}{a|}{89.36} &\multicolumn{1}{c|}{}\\
        \cdashline{3-10}
         &  & Nisioi et al.\ & Nisioi et al.\ & bleu(o,t) & SacreBLEU  & OTE & Nisioi et al.\ & this paper & \multicolumn{1}{a|}{88.10} &\multicolumn{1}{c|}{}\\
        \cdashline{3-10}
         &  & Nisioi et al.\ & Coop.\ \& Shard.\  & bleu(o,t) & Nisioi et al.\ & OTE & Nisioi et al.\ & Coop.\ \& Shard.\  & \textbf{80.75} &\multicolumn{1}{c|}{}\\
        \cdashline{3-10}
         &  & Nisioi et al.\ & this paper & bleu(o,t) & $\approx$Nisioi et al.\ & OTE & Nisioi et al.\ & this paper & \textbf{89.64} &\multicolumn{1}{c|}{}\\
        \cdashline{3-10}
         &  & Nisioi et al.\ & this paper & bleu(o,t) & SacreBLEU & OTE & Nisioi et al.\ & this paper & 88.80 &\multicolumn{1}{c|}{}\\
         \cline{2-12}
         & \multirow{4}{*}{SARI} & Nisioi et al.\ & Nisioi et al.\ & sari(o,s,t) & Nisioi et al.\ & OITE & Nisioi et al.\ &Nisioi et al.\ & \multicolumn{1}{g|}{\textbf{\textit{31.11}}} &\multicolumn{1}{c|}{\multirow{4}{*}{\textit{3.572}}}\\
        \cdashline{3-10}
         &  & Nisioi et al.\ & Nisioi et al.\ &  sari(o,s,t) & Nisioi et al.\ & OITE & Nisioi et al.\ & this paper & \multicolumn{1}{g|}{\textit{31.11}} &\multicolumn{1}{c|}{}\\
        \cdashline{3-10}
         &  & Nisioi et al.\ & Coop.\ \& Shard.\  & sari(o,s,t) & Nisioi et al.\ & OITE & Nisioi et al.\ & Coop.\ \& Shard.\  & \textbf{30.28} &\multicolumn{1}{c|}{}\\
        \cdashline{3-10}
         &  & Nisioi et al.\ & this paper & sari(o,s,t) & Nisioi et al.\ & OITE & Nisioi et al.\ & this paper & \textbf{29.12} &\multicolumn{1}{c|}{}\\
    \hline
\end{tabular}
\end{scriptsize}
    \caption{Conditions of measurement for each measurement carried out for the \textbf{NTS systems}. OTE = outputs vs.\ targets evaluation, OITE = outputs vs.\ inputs and targets evaluation. Shaded cells: evaluation of the same system outputs, i.e.\ the reproductions did not regenerate outputs. Bold: evaluation of (potentially) different system outputs, i.e.\ the reproductions did regenerate outputs.}
    \label{tab:nts-conditions} 
\end{table*}

If we conversely look just at those measurements (identifiable by boldfaced measured quantity values in Table~\ref{tab:nts-conditions}) where the reproducing team regenerated outputs (with the same system code) and evaluation scripts were the same, SARI CV$^*$ is 3.11 for the NTS\_def variants, and 4.05 for the NTS-w2v\_def variants (compared in both cases to 0 (perfect) when the same outputs are used). BLEU CV$^*$ is 2.154 for the NTS\_def variants (compared to 0.838 for same outputs but different evaluation scripts, as above), and 6.598 for the NTS-w2v\_def variants (compared to 1.314 for same outputs but different evaluation scripts). These 
differences arise simply from running the system in different environments.

The overall higher (worse) CV$^*$ values for NTS-w2v\_def variants (compared to NTS\_def) are likely to be partly due to the NTS models using one third party tool (openNMT), and the NTS-w2v models using two (openNMT and word2vec), i.e.\ the latter are more susceptible to changes in dependencies.

\subsection{QRA for PASS system}

The PASS system, developed by \citet{van2017pass}, generates football match reports from the perspective of each of the competing teams. The original study evaluated the system for Clarity, Fluency and Stance Identifiability in an evaluation with 20 evaluators and a test set of 10 output pairs. The evaluation was repeated with a slightly different evaluation interface and a different cohort of evaluators by \citet{mille:etal:2021}. Table~\ref{tab:pass-measurements} shows the results from the original and reproduction evaluations (columns~3 and~4), where the Clarity and Fluency results are the mean scores from 7-point agreement scales, and Identifiability results are the percentage of times the evaluators correctly guessed the team whose supporters a report was written for. Columns 6--9 show the corresponding sample size (number of reproductions plus original study), mean, standard deviation (stdev), the confidence interval (CI) for the standard deviation, and CV$^*$, all calculated on the shifted scores (see Section~\ref{ssec:formalisation}).  

Table~\ref{tab:pass-conditions} shows the values (here, paper identifiers) for the nine conditions of measurement introduced in Section~\ref{ssec:application}, for each of the six individual measurements (three evaluation measures times two studies). Note that both object conditions and the test set condition are the same, because \citeauthor{mille:etal:2021}\ used the system outputs shared by \citeauthor{van2017pass} The values for the \textit{Implemented by}, \textit{Procedure} and \textit{Performed by} conditions reflect the differences in the two evaluations in design, evaluator cohorts, and the teams that performed them. 

The scores vary to different degrees for the three measurands, with CV$^*$ lowest (reproducibility best) for Stance Identifiability, and highest (worst) for Fluency. These CV$^*$ results are likely to reflect that evaluators agreed more on Clarity than Fluency. Moreover, the binary stance identification assessment has better reproducibility than the other two criteria which are assessed on 7-point rating scales.

\subsection{QRA for essay scoring system variants}

\begin{table*}[ht]
   \centering
\setlength\tabcolsep{3.1pt} 
\renewcommand{\arraystretch}{1.25} 
\sffamily\selectfont
    \begin{scriptsize}
    \begin{tabular}{|l|c|c|c|c|c|c|c|c|c|c|c|c|c|c|}
    \hline
        & & \multicolumn{8}{c|}{Measured quantity value} &  &&&&\\
        \cline{3-10}
               &       & Vajjala & Huber \& & Arhiliuc & \multicolumn{3}{c|}{} & \multicolumn{2}{c|}{} & & & & & \\
        Object & Meas- & \& Rama & \multicolumn{1}{c|}{Coltekin} & \multicolumn{1}{c|}{ et al.} & \multicolumn{3}{c|}{Bestgen} & \multicolumn{2}{c|}{Caines \& Buttery} & Sample & mean & stdev & stdev 95\% CI & CV$^*$ $\downarrow$  \\
        \cline{3-10}
        &urand& seed 1 & seed 2 & seed ? & \multicolumn{2}{c|}{seed 1} & seed 2 & seed 1 & seed ? &size &&&&\\
        \cline{3-10}
        && e1 / i1 & e2 / i2 & e3 / i1 & e4 / i1 & e5 / i1 & e5 / i3 & e6 / i1 & e7 / i4 &&&&&\\
    \hline
    \hline
        mult-base & wF1 & 0.428 & 0.493 & 0.426 & 0.574 & 0.579 & 0.590 & 0.574 & 0.600 & 8 & 0.533  &  0.08  & [0.03, 0.12] &  14.633  \\
        mult-word$^-$ & wF1 & 0.721 & 0.603 & 0.605 & 0.606 & 0.720 & 0.732 & 0.606 & 0.740 & 8 & 0.667  &  0.07  & [0.03, 0.11] &  10.609 \\
        mult-word$^+$ & wF1 & 0.719 & 0.604 & 0.607 & 0.607 & 0.723 & 0.733 & 0.607 & 0.736 & 8 & 0.667  &  0.07  & [0.03, 0.11] &  10.440 \\ 
        mult-POS$^-$ & wF1 & 0.726 & 0.681 & 0.680 & 0.680 & 0.722 & 0.728 & 0.680 & 0.732 & 8 & 0.704  &  0.03  & [0.01, 0.04] &  3.818\\
        mult-POS$^+$ & wF1 & 0.724 & 0.680 & 0.680 & 0.681 & 0.725 & 0.729 & 0.681 & 0.731 & 8 & 0.704  &  0.03  & [0.01,  0.04] &  3.808 \\ 
        mult-dep$^-$ & wF1 & 0.703 & 0.660 & 0.650 & 0.651 & 0.699 & 0.711 & 0.651 & 0.710 & 8 & 0.679  &  0.03  & [0.01,  0.05] &  4.500 \\
        mult-dep$^+$ & wF1 & 0.693 & 0.661 & 0.652 & 0.653 & 0.699 & 0.712 & 0.653 & 0.716 & 8 & 0.68  &  0.03  & [0.01,  0.05] &  4.387\\ 
        mult-dom$^-$ & wF1 & 0.449 & 0.600 & 0.433 & 0.597 & 0.635 & 0.646 & 0.597 & 0.698 & 8 & 0.582  &  0.1  & [0.04,  0.15] &  17.147 \\
        mult-dom$^+$ & wF1 & 0.471 & 0.647 & 0.447 & 0.647 & 0.696 & 0.711 & 0.647 & 0.726 & 8 & 0.624  &  0.11  & [0.05,  0.18] &  18.248 \\ 
        mult-emb$^-$ & wF1 & 0.693 & 0.658 & 0.683 & 0.668 & 0.692 & 0.689 & 0.659 & 0.391 & 8 & 0.642  &  0.11  & [0.04,  0.17] &  17.033 \\
        mult-emb$^+$ & wF1 & 0.689 & 0.662 & 0.681 & 0.659 & 0.681 & 0.684 & 0.657 & 0.401 & 8 & 0.639  &  0.1  & [0.04,  0.16] &  16.226 \\ 
    \hline
    \end{tabular}
    \end{scriptsize} 
    \caption{Precision (CV$^*$) and component measures (mean, standard deviation, standard deviation confidence intervals) for measured quantity values obtained in multiple measurements of the \textbf{essay scoring systems}. Seed $i$ = different approaches to random seeding and cross-validation; e$i$ = different compile/run-time environments; i$i$ = different test data sets and/or cross-validation folds. }
    \label{tab:estc-measurements}
\end{table*}

The 11 multilingual essay scoring system variants reported by \citet{vajjala-rama-2018-experiments} were evaluated by weighted F1 (wF1) score.
Table~\ref{tab:estc-measurements} shows wF1 scores for the 11 multilingual system variants from each of the five papers, alongside the 11 corresponding CV$^*$ values. Table~\ref{tab:estc-conditions} in the appendix shows the corresponding conditions of measurement.
The baseline classifier (mult-base) uses document length (number of words) as its only feature. For the other variants, +/- indicates that the multilingual classifier was / was not given information about which language the input was in; the mult-word variants use word n-grams only; mult-word uses POS (part of speech) tag n-grams only; mult-dep uses n-grams over dependency relation, dependent POS, and head POS triples; mult-dom uses domain-specific linguistic features including document length, lexical richness and errors; mult-emb uses word and character embeddings. The mult-base and mult-dom models are logistic regressors, the others are random forests.

A very clear picture emerges: system variant pairs that differ only in whether they do or do not use language information have very similar CV scores. For example, mult-POS$^-$ (POS n-grams without language information) and mult-POS$^+$ (POS n-grams with language information) both have a very good degree of wF1-reproducibility, their CV$^*$ being 3.818 and 3.808 respectively; mult-word$^-$ (word n-grams without language information) and mult-word$^+$ (word n-grams with language information) have notably higher CV$^*$, around 10. 
This tendency holds for all such pairs, indicating that using language information makes next to no difference to reproducibility. Moreover, the mult-dom and mult-emb variants all have similar CV$^*$.\footnote{The high CV$^*$ for the baseline system may be due to an issue wiith the evaluation code (macro-F1 instead of weighted-F1), as reported by \citeauthor{bestgen-2020-reproducing} (Section 3.2, first paragraph), \citeauthor{caines-buttery-2020-reprolang} (Section 2.5, one before last paragraph) and \citeauthor{huber-coltekin-2020-reproduction} (Section 3.2, second paragraph).}

The indication is that the syntactic information is obtained/used in a way that is particularly reproducible, whereas the domain-specific information and the embeddings are obtained/used in a way that is particularly hard to reproduce. Overall, the random forest models using syntactic features have the best reproducibility; the logistic regressors using domain-specific features have the worst.

\section{Discussion}\label{sec:discussion}

Quantified reproducibility assessment (QRA) enables assessment of the {degree} of reproducibility of evaluation results for any given system and evaluation measure in a way that is scale-invariant\footnote{If evaluation scores are multiplied by a common factor, CV$^*$ does not change.} and comparable across different QRAs, for reproductions involving either the same or different original studies. Moreover, formally capturing (dis)similarities between systems and evaluation designs enables reproducibility to be assessed relative to such (dis)similarities.
{In combination, a set of results from QRA tests for the same system and evaluation measure can provide pointers to which aspects of the system and evaluation might be associated with low reproducibility. E.g.\ for the wF1 evaluations of the essay scoring systems above, it is clear that variations in reproducibility are associated at least in part with the different features used by systems.}

It might be expected that the reproducibility of human-assessed evaluations is generally worse than metric-assessed. Our study revealed a more mixed picture. As expected, the Fluency and Clarity evaluations of the PASS system were among those with highest CV$^*$, and the BLEU and SARI evaluation of the NTS systems and wF1 evaluation of the mult-POS and mult-dep systems were among those with lowest CV$^*$. However, human-assessed Stance Identifiability of PASS was among the most reproducible, and metric-assessed wF1 of mult-base, mult-dom and mult-emb were among the worst.

In this paper, our focus has been QRA testing of existing research results. However, ideally, QRA would be built into new method development from the outset, where at first reporting, a detailed standardised set of conditions of measurement is specified, and repeatability tests (where all conditions are identical except for the team conducting the tests, see Section~\ref{ssec:formalisation}) are performed to determine baseline reproducibility. Such repeatability QRA would provide quality assurance for new methods as well as important pointers for future reproductions regarding what degree of reproducibility to expect for given (types of) methods. 

If this is not possible, post-hoc reproducibility QRA (where there are differences in conditions of measurement values) is performed instead. If this yields high (poor) CV$^*$, one way to proceed is to minimise differences in conditions of measurement between the studies and observe the effect on CV$^*$, changing aspects of system and evaluation design and adding further conditions of measurement if need be. For human evaluation in particular, persistently high CV$^*$ would indicate a problem with the method itself.

\vspace{-.1cm}
\section{Conclusion}\label{sec:conclusion}
\vspace{-.1cm}

We have described an approach to quantified reproducibility assessment (QRA) based on concepts and definitions from metrology, and tested it on 18 system and evaluation measure combinations involving diverse NLP tasks and types of evaluation. 

QRA produces a single score that quantifies the degree of reproducibility of a given system and evaluation measure, on the basis of the scores from, and differences between, multiple reproductions of the same original study. We found that the approach facilitates insights into sources of variation between reproductions, produces results that are comparable across different reproducibility assessments, and provides pointers about what needs to be changed in system and/or evaluation design to improve reproducibility.

A recent survey \cite{belz2021systematic} found that just 14\% of the 513 original/reproduction score pairs analysed were exactly
the same. Judging the remainder simply `not reproduced’ is of limited usefulness, as some are
much closer to being the same than others. At the same time,
assessments of whether the same conclusions can
be drawn on the basis of different scores involve subjective judgments and are prone to disagreement among assessors.
Quantifying the closeness of results as in QRA, and, over time, establishing expected levels of closeness, seems a better
way forward.

\vspace{-.1cm}
\section*{Acknowledgements}
\vspace{-.1cm}

We are grateful to the anonymous reviewers and area chairs for their exceptionally detailed and helpful feedback.

Popovi\'c's work on this s study was funded by the 
ADAPT SFI Centre for Digital Media Technology
which is funded by Science Foundation Ireland
through the SFI Research Centres Programme, and
co-funded under the European Regional Development Fund (ERDF) through Grant 13/RC/2106. Mille’s work was supported by
the European Commission under the H2020 program contract numbers 786731, 825079, 870930
and 952133.

\bibliography{anthology,repro}

\begin{thebibliography}{28}
\expandafter\ifx\csname natexlab\endcsname\relax\def\natexlab#1{#1}\fi

\bibitem[{Ahmed(1995)}]{ahmed1995pooling}
SE~Ahmed. 1995.
\newblock A pooling methodology for coefficient of variation.
\newblock \emph{Sankhy{\=a}: The Indian Journal of Statistics, Series B}, pages
  57--75.

\bibitem[{Arhiliuc et~al.(2020)Arhiliuc, Mitrovi{\'c}, and
  Granitzer}]{arhiliuc-etal-2020-language}
Cristina Arhiliuc, Jelena Mitrovi{\'c}, and Michael Granitzer. 2020.
\newblock \href {https://www.aclweb.org/anthology/2020.lrec-1.690} {Language
  proficiency scoring}.
\newblock In \emph{Proceedings of The 12th Language Resources and Evaluation
  Conference}, pages 5624--5630, Marseille, France. European Language Resources
  Association.

\bibitem[{{Association for Computing Machinery}(2020)}]{acm2020artifact}
{Association for Computing Machinery}. 2020.
\newblock \href
  {https://www.acm.org/publications/policies/artifact-review-and-badging-current}
  {Artifact review and badging {V}ersion 1.1}.
\newblock Accessed {A}ugust 24, 2020.

\bibitem[{Belz et~al.(2021{\natexlab{a}})Belz, Agarwal, Shimorina, and
  Reiter}]{belz2021systematic}
Anya Belz, Shubham Agarwal, Anastasia Shimorina, and Ehud Reiter.
  2021{\natexlab{a}}.
\newblock A systematic review of reproducibility research in natural language
  processing.
\newblock In \emph{Proceedings of the 16th Conference of the European Chapter
  of the Association for Computational Linguistics: Main Volume}, pages
  381--393.

\bibitem[{Belz et~al.(2021{\natexlab{b}})Belz, Shimorina, Agarwal, and
  Reiter}]{belz2021reprogen}
Anya Belz, Anastasia Shimorina, Shubham Agarwal, and Ehud Reiter.
  2021{\natexlab{b}}.
\newblock The reprogen shared task on reproducibility of human evaluations in
  {NLG}: {O}verview and results.
\newblock In \emph{The 14th International Conference on Natural Language
  Generation}.

\bibitem[{Bestgen(2020)}]{bestgen-2020-reproducing}
Yves Bestgen. 2020.
\newblock \href {https://www.aclweb.org/anthology/2020.lrec-1.687} {Reproducing
  monolingual, multilingual and cross-lingual {CEFR} predictions}.
\newblock In \emph{Proceedings of The 12th Language Resources and Evaluation
  Conference}, pages 5595--5602, Marseille, France. European Language Resources
  Association.

\bibitem[{Branco et~al.(2020)Branco, Calzolari, Vossen, Van~Noord, van
  Uytvanck, Silva, Gomes, Moreira, and Elbers}]{branco-etal-2020-shared}
Ant\'onio Branco, Nicoletta Calzolari, Piek Vossen, Gertjan Van~Noord, Dieter
  van Uytvanck, Jo\~ao Silva, Lu\'is Gomes, Andr\'e Moreira, and Willem Elbers.
  2020.
\newblock \href {https://www.aclweb.org/anthology/2020.lrec-1.680} {A shared
  task of a new, collaborative type to foster reproducibility: A first exercise
  in the area of language science and technology with {REPROLANG}2020}.
\newblock In \emph{Proceedings of The 12th Language Resources and Evaluation
  Conference}, pages 5539--5545, Marseille, France. European Language Resources
  Association.

\bibitem[{Caines and Buttery(2020)}]{caines-buttery-2020-reprolang}
Andrew Caines and Paula Buttery. 2020.
\newblock \href {https://www.aclweb.org/anthology/2020.lrec-1.689} {{REPROLANG}
  2020: Automatic proficiency scoring of {C}zech, {E}nglish, {G}erman,
  {I}talian, and {S}panish learner essays}.
\newblock In \emph{Proceedings of The 12th Language Resources and Evaluation
  Conference}, pages 5614--5623, Marseille, France. European Language Resources
  Association.

\bibitem[{Cooper and Shardlow(2020)}]{cooper-shardlow-2020-combinmt}
Michael Cooper and Matthew Shardlow. 2020.
\newblock \href {https://www.aclweb.org/anthology/2020.lrec-1.686}
  {{C}ombi{NMT}: An exploration into neural text simplification models}.
\newblock In \emph{Proceedings of The 12th Language Resources and Evaluation
  Conference}, pages 5588--5594, Marseille, France. European Language Resources
  Association.

\bibitem[{Drummond(2009)}]{drummond2009replicability}
Chris Drummond. 2009.
\newblock \href {http://cogprints.org/7691/7/ICMLws09.pdf} {Replicability is
  not reproducibility: nor is it good science}.
\newblock Presented at 4th Workshop on Evaluation Methods for Machine Learning
  held at ICML'09.

\bibitem[{Huber and
  {\c{C}}{\"o}ltekin(2020)}]{huber-coltekin-2020-reproduction}
Eva Huber and {\c{C}}a{\u{g}}r{\i} {\c{C}}{\"o}ltekin. 2020.
\newblock \href {https://www.aclweb.org/anthology/2020.lrec-1.688}
  {Reproduction and replication: A case study with automatic essay scoring}.
\newblock In \emph{Proceedings of The 12th Language Resources and Evaluation
  Conference}, pages 5603--5613, Marseille, France. European Language Resources
  Association.

\bibitem[{JCGM(2012)}]{jcgm2012international}
JCGM. 2012.
\newblock \href
  {https://www.bipm.org/utils/common/documents/jcgm/JCGM_200_2012.pdf}
  {International vocabulary of metrology: {B}asic and general concepts and
  associated terms ({VIM})}.
\newblock Joint Committee for Guides in Metrology,
  \url{https://www.bipm.org/utils/common/documents/jcgm/JCGM_200_2012.pdf}.

\bibitem[{Mille et~al.(2021)Mille, Ferreira, Belz, and Davis}]{mille:etal:2021}
Simon Mille, Thiago~Castro Ferreira, Anya Belz, and Brian Davis. 2021.
\newblock Another {PASS}: {A} reproduction study of the human evaluation of a
  football report generation system.
\newblock In \emph{Proceedings of the 14th International Conference on Natural
  Language Generation (INLG 2021)}.

\bibitem[{Nisioi et~al.(2017)Nisioi, {\v{S}}tajner, Ponzetto, and
  Dinu}]{nisioi-etal-2017-exploring}
Sergiu Nisioi, Sanja {\v{S}}tajner, Simone~Paolo Ponzetto, and Liviu~P. Dinu.
  2017.
\newblock \href {https://doi.org/10.18653/v1/P17-2014} {Exploring neural text
  simplification models}.
\newblock In \emph{Proceedings of the 55th Annual Meeting of the Association
  for Computational Linguistics (Volume 2: Short Papers)}, pages 85--91,
  Vancouver, Canada. Association for Computational Linguistics.

\bibitem[{Pedersen(2008)}]{pedersen2008empiricism}
Ted Pedersen. 2008.
\newblock Empiricism is not a matter of faith.
\newblock \emph{Computational Linguistics}, 34(3):465--470.

\bibitem[{Pineau(2020)}]{pineau2020checklist}
Joelle Pineau. 2020.
\newblock \href
  {https://www.cs.mcgill.ca/~jpineau/ReproducibilityChecklist.pdf} {The machine
  learning reproducibility checklist v2.0}.

\bibitem[{Popovi{\'c} and Belz(2021)}]{popovic-belz-2021-reproduction}
Maja Popovi{\'c} and Anya Belz. 2021.
\newblock \href {https://aclanthology.org/2021.inlg-1.31} {A reproduction study
  of an annotation-based human evaluation of {MT} outputs}.
\newblock In \emph{Proceedings of the 14th International Conference on Natural
  Language Generation}, pages 293--300, Aberdeen, Scotland, UK. Association for
  Computational Linguistics.

\bibitem[{Post(2018)}]{post2018call}
Matt Post. 2018.
\newblock A call for clarity in reporting bleu scores.
\newblock \emph{WMT 2018}, page 186.

\bibitem[{Rao(1973)}]{rao1973linear}
Calyampudi~Radhakrishna Rao. 1973.
\newblock \emph{Linear statistical inference and its applications}.
\newblock Wiley.

\bibitem[{Rougier et~al.(2017)Rougier, Hinsen, Alexandre, Arildsen, Barba,
  Benureau, Brown, De~Buyl, Caglayan, Davison et~al.}]{rougier2017sustainable}
Nicolas~P. Rougier, Konrad Hinsen, Fr{\'e}d{\'e}ric Alexandre, Thomas Arildsen,
  Lorena~A Barba, Fabien~CY Benureau, C~Titus Brown, Pierre De~Buyl, Ozan
  Caglayan, Andrew~P Davison, et~al. 2017.
\newblock Sustainable computational science: {T}he {ReScience} initiative.
\newblock \emph{PeerJ Computer Science}, 3:e142.

\bibitem[{Schloss(2018)}]{schloss2018identifying}
Patrick~D. Schloss. 2018.
\newblock Identifying and overcoming threats to reproducibility, replicability,
  robustness, and generalizability in microbiome research.
\newblock \emph{MBio}, 9(3).

\bibitem[{Shimorina and Belz(2021)}]{shimorina2021human}
Anastasia Shimorina and Anya Belz. 2021.
\newblock The human evaluation datasheet 1.0: {A} template for recording
  details of human evaluation experiments in {NLP}.
\newblock \emph{arXiv preprint arXiv:3910940}.

\bibitem[{Sokal and Rohlf(1971)}]{sokal:rohlf:1969}
R.R. Sokal and F.J. Rohlf. 1971.
\newblock \emph{Biometry: The Principles and Practice of Statistics in
  Biological Research}.
\newblock WH Freeman.

\bibitem[{Vajjala and Rama(2018)}]{vajjala-rama-2018-experiments}
Sowmya Vajjala and Taraka Rama. 2018.
\newblock \href {https://doi.org/10.18653/v1/W18-0515} {Experiments with
  universal {CEFR} classification}.
\newblock In \emph{Proceedings of the Thirteenth Workshop on Innovative Use of
  {NLP} for Building Educational Applications}, pages 147--153, New Orleans,
  Louisiana. Association for Computational Linguistics.

\bibitem[{van~der Lee et~al.(2017)van~der Lee, Krahmer, and
  Wubben}]{van2017pass}
Chris van~der Lee, Emiel Krahmer, and Sander Wubben. 2017.
\newblock {PASS}: {A} {D}utch data-to-text system for soccer, targeted towards
  specific audiences.
\newblock In \emph{Proceedings of the 10th International Conference on Natural
  Language Generation}, pages 95--104.

\bibitem[{Whitaker(2017)}]{whitaker2017}
Kirstie Whitaker. 2017.
\newblock The {MT} {R}eproducibility {C}hecklist.
\newblock
  \url{{https://www.cs.mcgill.ca/~jpineau/ReproducibilityChecklist.pdf}}.

\bibitem[{Wieling et~al.(2018)Wieling, Rawee, and van
  Noord}]{wieling2018reproducibility}
Martijn Wieling, Josine Rawee, and Gertjan van Noord. 2018.
\newblock Reproducibility in computational linguistics: Are we willing to
  share?
\newblock \emph{Computational Linguistics}, 44(4):641--649.

\bibitem[{Xu et~al.(2016)Xu, Napoles, Pavlick, Chen, and
  Callison-Burch}]{xu2016optimizing}
Wei Xu, Courtney Napoles, Ellie Pavlick, Quanze Chen, and Chris Callison-Burch.
  2016.
\newblock Optimizing statistical machine translation for text simplification.
\newblock \emph{Transactions of the Association for Computational Linguistics},
  4:401--415.

\end{thebibliography}
\bibliographystyle{acl_natbib}

\appendix

\onecolumn

\section{Conditions of Measurement for the Essay Scoring Systems}
\label{sec:appendix}
Table~\ref{tab:estc-conditions2} shows the conditions of measurement for each of the 88 individual measurements for the Essay Scoring Systems.

\begin{table*}[b!]
    \centering
\setlength\tabcolsep{2.75pt} 
\renewcommand{\arraystretch}{1.25} 
\sffamily\selectfont
\begin{scriptsize}
\begin{tabular}{|c|c|c|c|c|c|c|c|c|c|c|c|c|}
    \hline
     & & \multicolumn{2}{c|}{\multirow{2}{*}{\textit{Object conditions}}} & \multicolumn{2}{c|}{{\textit{Measurement method}}} & \multicolumn{3}{c|}{{\textit{Measurement procedure}}} & \multicolumn{1}{c|}{{Measured}} & \multicolumn{1}{c|}{}\\
    Object & Measurand & \multicolumn{2}{c|}{} & \multicolumn{2}{c|}{{\textit{conditions}}} & \multicolumn{3}{c|}{{\textit{conditions}}} & \multicolumn{1}{c|}{{quantity}} & \multicolumn{1}{c|}{\textit{CV$^*$}} \\
    \cline{3-9}
     &  & {\textit{Code by}} & {\textit{Comp./trained by}} & \textit{Method} & \textit{Implem.\ by } & \textit{Procedure} & \multicolumn{1}{c|}{\textit{Test set}} & \multicolumn{1}{c|}{\textit{Performed by}}  & {value} & \multicolumn{1}{c|}{}\\
    \hline
    \hline
        \multirow{8}{*}{mult-base} & \multirow{8}{*}{wF1} & Va.\& Ra. & Va.\& Ra. & wF1(o,t) & Va.\& Ra.& OTE & Va.\& Ra.& Va.\& Ra.& 0.428 & \multicolumn{1}{c|}{\multirow{8}{*}{\textit{14.633}}}\\
        \cdashline{3-10}
         &  & Va.\& Ra. & Huber \& Coltekin& wF1(o,t) & Va.\& Ra.& OTE & Va.\& Ra.& Huber \& Coltekin& 0.493 & \\
        \cdashline{3-10}
         &  & Va.\& Ra. & Arhiliuc et al. & wF1(o,t) & Va.\& Ra.& OTE & Va.\& Ra.& Arhiliuc et al.& 0.426 & \\
        \cdashline{3-10}
         &  & Va.\& Ra. & Va.\& Ra. & wF1(o,t) &  Va.\& Ra. & OTE &  Va.\& Ra. & Bestgen & 0.574 & \\
        \cdashline{3-10}
         &  & Va.\& Ra. & Va.\& Ra. & wF1(o,t) &  Va.\& Ra. & OTE &  Va.\& Ra. & Bestgen & 0.579 & \\
        \cdashline{3-10}
         &  & Va.\& Ra. & Va.\& Ra. & wF1(o,t) &  $\approx$Va.\& Ra. & OTE &  Va.\& Ra. & Bestgen & 0.590 & \\
        \cdashline{3-10}
         &  & Va.\& Ra. & Va.\& Ra. & wF1(o,t) &  Va.\& Ra. & OTE &  Va.\& Ra. & Cai. \& But. & 0.574 & \\
        \cdashline{3-10}
         &  & Cai. \& But. & Cai. \& But. & wF1(o,t) & Cai. \& But. & OTE &  Va.\&Ra. & Cai. \& But. & 0.600 & \\
        \hline
        \multirow{8}{*}{mult-word$^-$} & \multirow{8}{*}{wF1} & Va.\& Ra. & Va.\& Ra. & wF1(o,t) & Va.\& Ra.& OTE & Va.\& Ra.& Va.\& Ra.& 0.721 & \multicolumn{1}{c|}{\multirow{8}{*}{\textit{10.609}}}\\
        \cdashline{3-10}
         &  & Va.\& Ra. & Huber \& Coltekin& wF1(o,t) & Va.\& Ra.& OTE & Va.\& Ra.& Huber \& Coltekin& 0.603 & \\
        \cdashline{3-10}
         &  & Va.\& Ra. & Arhiliuc et al. & wF1(o,t) & Va.\& Ra.& OTE & Va.\& Ra.& Arhiliuc et al.& 0.605   & \\
        \cdashline{3-10}
         &  & Va.\& Ra. & Va.\& Ra. & wF1(o,t) &  Va.\& Ra. & OTE &  Va.\& Ra. & Bestgen & 0.606 & \\
        \cdashline{3-10}
        &  & Va.\& Ra. & Va.\& Ra. & wF1(o,t) &  Va.\& Ra. & OTE &  Va.\& Ra. & Bestgen & 0.720 & \\
        \cdashline{3-10}
         &  & Va.\& Ra. & Va.\& Ra. & wF1(o,t) &  $\approx$Va.\& Ra. & OTE &  Va.\& Ra. & Bestgen & 0.732 & \\
        \cdashline{3-10}
         &  & Va.\& Ra. & Va.\& Ra. & wF1(o,t) &  Va.\& Ra. & OTE &  Va.\& Ra. & Cai. \& But. & 0.606 & \\
        \cdashline{3-10}
         &  & Cai. \& But. & Cai. \& But. & wF1(o,t) & Cai. \& But. & OTE &  Va.\&Ra. & Cai. \& But. & 0.740 & \\
        \hline
        \multirow{8}{*}{mult-word$^+$} & \multirow{8}{*}{wF1} & Va.\& Ra. & Va.\& Ra. & wF1(o,t) & Va.\& Ra.& OTE & Va.\& Ra.& Va.\& Ra.& 0.719 & \multicolumn{1}{c|}{\multirow{8}{*}{\textit{10.44}}}\\
        \cdashline{3-10}
         &  & Va.\& Ra. & Huber \& Coltekin& wF1(o,t) & Va.\& Ra.& OTE & Va.\& Ra.& Huber \& Coltekin& 0.604 & \\
        \cdashline{3-10}
         &  & Va.\& Ra. & Arhiliuc et al. & wF1(o,t) & Va.\& Ra.& OTE & Va.\& Ra.& Arhiliuc et al.& 0.607 & \\
        \cdashline{3-10}
         &  & Va.\& Ra. & Va.\& Ra. & wF1(o,t) &  Va.\& Ra. & OTE &  Va.\& Ra. & Bestgen & 0.607 &\\
        \cdashline{3-10}
         &  & Va.\& Ra. & Va.\& Ra. & wF1(o,t) &  Va.\& Ra. & OTE &  Va.\& Ra. & Bestgen & 0.723 &\\
        \cdashline{3-10}
         &  & Va.\& Ra. & Va.\& Ra. & wF1(o,t) &  $\approx$Va.\& Ra. & OTE &  Va.\& Ra. & Bestgen & 0.733 & \\
        \cdashline{3-10}
         &  & Va.\& Ra. & Va.\& Ra. & wF1(o,t) &  Va.\& Ra. & OTE &  Va.\& Ra. & Cai. \& But. & 0.607 & \\
        \cdashline{3-10}
         &  & Cai. \& But. & Cai. \& But. & wF1(o,t) & Cai. \& But. & OTE &  Va.\&Ra. & Cai. \& But. & 0.736 & \\
        \hline
        \multirow{8}{*}{mult-POS$^-$} & \multirow{8}{*}{wF1} & Va.\& Ra. & Va.\& Ra. & wF1(o,t) & Va.\& Ra.& OTE & Va.\& Ra.& Va.\& Ra.& 0.726 &  \multicolumn{1}{c|}{\multirow{8}{*}{\textit{3.818}}}\\
        \cdashline{3-10}
         &  & Va.\& Ra. & Huber \& Coltekin& wF1(o,t) & Va.\& Ra.& OTE & Va.\& Ra.& Huber \& Coltekin& 0.681 &\\
        \cdashline{3-10}
         &  & Va.\& Ra.& Arhiliuc et al.& wF1(o,t) & Va.\& Ra.& OTE & Va.\& Ra.& Arhiliuc et al.& 0.680 &\\
        \cdashline{3-10}
        &  & Va.\& Ra. & Va.\& Ra. & wF1(o,t) &  Va.\& Ra. & OTE &  Va.\& Ra. & Bestgen & 0.680 &\\
        \cdashline{3-10}
         &  & Va.\& Ra. & Va.\& Ra. & wF1(o,t) &  Va.\& Ra. & OTE &  Va.\& Ra. & Bestgen & 0.722 &\\
        \cdashline{3-10}
         &  & Va.\& Ra. & Va.\& Ra. & wF1(o,t) &  $\approx$Va.\& Ra. & OTE &  Va.\& Ra. & Bestgen & 0.728 &\\
        \cdashline{3-10}
         &  & Va.\& Ra. & Va.\& Ra. & wF1(o,t) &  Va.\& Ra. & OTE &  Va.\& Ra. & Cai. \& But. & 0.680 &\\
        \cdashline{3-10}
         &  & Cai. \& But. & Cai. \& But. & wF1(o,t) & Cai. \& But. & OTE &  Va.\&Ra. & Cai. \& But. & 0.732 &\\
        \hline
        \multirow{8}{*}{mult-POS$^+$} & \multirow{8}{*}{wF1} & Va.\& Ra. & Va.\& Ra. & wF1(o,t) & Va.\& Ra.& OTE & Va.\& Ra.& Va.\& Ra.& 0.724 &  \multicolumn{1}{c|}{\multirow{8}{*}{\textit{3.808}}}\\
        \cdashline{3-10}
         &  & Va.\& Ra. & Huber \& Coltekin& wF1(o,t) & Va.\& Ra.& OTE & Va.\& Ra.& Huber \& Coltekin & 0.680 &\\
        \cdashline{3-10}
         &  & Va.\& Ra.& Arhiliuc et al.& wF1(o,t) & Va.\& Ra.& OTE & Va.\& Ra.& Arhiliuc et al.& 0.680 &\\
        \cdashline{3-10}
         &  & Va.\& Ra. & Va.\& Ra. & wF1(o,t) &  Va.\& Ra. & OTE &  Va.\& Ra. & Bestgen & 0.681 &\\
        \cdashline{3-10}
         &  & Va.\& Ra. & Va.\& Ra. & wF1(o,t) &  Va.\& Ra. & OTE &  Va.\& Ra. & Bestgen & 0.725 &\\
        \cdashline{3-10}
         &  & Va.\& Ra. & Va.\& Ra. & wF1(o,t) &  $\approx$Va.\& Ra. & OTE &  Va.\& Ra. & Bestgen & 0.729 &\\
        \cdashline{3-10}
        &  & Va.\& Ra. & Va.\& Ra. & wF1(o,t) &  Va.\& Ra. & OTE &  Va.\& Ra. & Cai. \& But. & 0.681 &\\
        \cdashline{3-10}
         &  & Cai. \& But. & Cai. \& But. & wF1(o,t) & Cai. \& But. & OTE &  Va.\&Ra. & Cai. \& But. & 0.731 &\\
        \hline
        \multirow{8}{*}{mult-dep$^-$} & \multirow{8}{*}{wF1} & Va.\& Ra. & Va.\& Ra. & wF1(o,t) & Va.\& Ra.& OTE & Va.\& Ra.& Va.\& Ra.& 0.703 & \multicolumn{1}{c|}{\multirow{8}{*}{\textit{ 4.5}}}\\
        \cdashline{3-10}
         &  & Va.\& Ra. & Huber \& Coltekin& wF1(o,t) & Va.\& Ra.& OTE & Va.\& Ra.& Huber \& Coltekin& 0.660 & \\
        \cdashline{3-10}
         &  & Va.\& Ra.& Arhiliuc et al.& wF1(o,t) & Va.\& Ra.& OTE & Va.\& Ra.& Arhiliuc et al.& 0.650 & \\
        \cdashline{3-10}
         &  & Va.\& Ra. & Va.\& Ra. & wF1(o,t) &  Va.\& Ra. & OTE &  Va.\& Ra. & Bestgen & 0.651 & \\
        \cdashline{3-10}
         &  & Va.\& Ra. & Va.\& Ra. & wF1(o,t) &  Va.\& Ra. & OTE &  Va.\& Ra. & Bestgen & 0.699 & \\
        \cdashline{3-10}
         &  & Va.\& Ra. & Va.\& Ra. & wF1(o,t) &  $\approx$Va.\& Ra. & OTE &  Va.\& Ra. & Bestgen & 0.711 & \\
        \cdashline{3-10}
         &  & Va.\& Ra. & Va.\& Ra. & wF1(o,t) &  Va.\& Ra. & OTE &  Va.\& Ra. & Cai. \& But. & 0.651 & \\
        \cdashline{3-10}
         &  & Cai. \& But. & Cai. \& But. & wF1(o,t) & Cai. \& But. & OTE &  Va.\&Ra. & Cai. \& But. & 0.710 & \\
        \hline
        \multirow{8}{*}{mult-dep$^+$} & \multirow{8}{*}{wF1} & Va.\& Ra. & Va.\& Ra. & wF1(o,t) & Va.\& Ra.& OTE & Va.\& Ra.& Va.\& Ra.& 0.693 & \multicolumn{1}{c|}{\multirow{8}{*}{\textit{4.387}}}\\
        \cdashline{3-10}
         &  & Va.\& Ra. & Huber \& Coltekin& wF1(o,t) & Va.\& Ra.& OTE & Va.\& Ra.& Huber \& Coltekin& 0.661 & \\
        \cdashline{3-10}
         &  & Va.\& Ra.& Arhiliuc et al.& wF1(o,t) & Va.\& Ra.& OTE & Va.\& Ra.& Arhiliuc et al.& 0.652 & \\
        \cdashline{3-10}
         &  & Va.\& Ra. & Va.\& Ra. & wF1(o,t) &  Va.\& Ra. & OTE &  Va.\& Ra. & Bestgen & 0.653 & \\
        \cdashline{3-10}
         &  & Va.\& Ra. & Va.\& Ra. & wF1(o,t) &  Va.\& Ra. & OTE &  Va.\& Ra. & Bestgen & 0.699 & \\
        \cdashline{3-10}
         &  & Va.\& Ra. & Va.\& Ra. & wF1(o,t) &  $\approx$Va.\& Ra. & OTE &  Va.\& Ra. & Bestgen & 0.712 & \\
        \cdashline{3-10}
         &  & Va.\& Ra. & Va.\& Ra. & wF1(o,t) &  Va.\& Ra. & OTE &  Va.\& Ra. & Cai. \& But. & 0.653 & \\
        \cdashline{3-10}
         &  & Cai. \& But. & Cai. \& But. & wF1(o,t) & Cai. \& But. & OTE &  Va.\&Ra. & Cai. \& But. & 0.716 & \\
        \hline
        \multicolumn{11}{l}{}\\
        \multicolumn{11}{l}{\textit{Table continued on next page.}}\\
\end{tabular}
\end{scriptsize}
\end{table*}

\begin{table*}[t!]
    \centering
\setlength\tabcolsep{2.75pt} 
\renewcommand{\arraystretch}{1.25} 
\sffamily\selectfont
\begin{scriptsize}
\begin{tabular}{|c|c|c|c|c|c|c|c|c|c|c|c|c|}
    \hline
     & & \multicolumn{2}{c|}{\multirow{2}{*}{\textit{Object conditions}}} & \multicolumn{2}{c|}{{\textit{Measurement method}}} & \multicolumn{3}{c|}{{\textit{Measurement procedure}}} & \multicolumn{1}{c|}{{Measured}} & \multicolumn{1}{c|}{}\\
    Object & Measurand & \multicolumn{2}{c|}{} & \multicolumn{2}{c|}{{\textit{conditions}}} & \multicolumn{3}{c|}{{\textit{conditions}}} & \multicolumn{1}{c|}{{quantity}} & \multicolumn{1}{c|}{\textit{CV$^*$}} \\
    \cline{3-9}
     &  & {\textit{Code by}} & {\textit{Comp./trained by}} & \textit{Method} & \textit{Implem.\ by } & \textit{Procedure} & \multicolumn{1}{c|}{\textit{Test set}} & \multicolumn{1}{c|}{\textit{Performed by}}  & {value} & \multicolumn{1}{c|}{}\\
    \hline
    \hline
        \multirow{8}{*}{mult-dom$^-$} & \multirow{8}{*}{wF1} & Va.\& Ra. & Va.\& Ra. & wF1(o,t) & Va.\& Ra.& OTE & Va.\& Ra.& Va.\& Ra.& 0.449 & \multicolumn{1}{c|}{\multirow{8}{*}{\textit{17.147}}}\\
        \cdashline{3-10}
         &  & Va.\& Ra. & Huber \& Coltekin& wF1(o,t) & Va.\& Ra.& OTE & Va.\& Ra.& Huber \& Coltekin& 0.600 & \\
        \cdashline{3-10}
         &  & Va.\& Ra.& Arhiliuc et al.& wF1(o,t) & Va.\& Ra. & OTE & Va.\& Ra.& Arhiliuc et al.& 0.433 & \\
        \cdashline{3-10}
         &  & Va.\& Ra. & Va.\& Ra. & wF1(o,t) &  Va.\& Ra. & OTE &  Va.\& Ra. & Bestgen & 0.597 & \\
        \cdashline{3-10}
         &  & Va.\& Ra. & Va.\& Ra. & wF1(o,t) &  Va.\& Ra. & OTE &  Va.\& Ra. & Bestgen & 0.635 & \\
        \cdashline{3-10}
         &  & Va.\& Ra. & Va.\& Ra. & wF1(o,t) &  $\approx$Va.\& Ra. & OTE &  Va.\& Ra. & Bestgen & 0.646 & \\
        \cdashline{3-10}
         &  & Va.\& Ra. & Va.\& Ra. & wF1(o,t) &  Va.\& Ra. & OTE &  Va.\& Ra. & Cai. \& But. & 0.597 & \\
        \cdashline{3-10}
         &  & Cai. \& But. & Cai. \& But. & wF1(o,t) & Cai. \& But. & OTE &  Va.\&Ra. & Cai. \& But. & 0.698 & \\
        \hline
        \multirow{8}{*}{mult-dom$^+$} & \multirow{8}{*}{wF1} & Va.\& Ra. & Va.\& Ra. & wF1(o,t) & Va.\& Ra.& OTE & Va.\& Ra.& Va.\& Ra.& 0.471 & \multicolumn{1}{c|}{\multirow{8}{*}{\textit{18.248}}}\\
        \cdashline{3-10}
         &  & Va.\& Ra. & Huber \& Coltekin& wF1(o,t) & Va.\& Ra.& OTE & Va.\& Ra.& Huber \& Coltekin& 0.647 & \\
        \cdashline{3-10}
         &  & Va.\& Ra.& Arhiliuc et al..& wF1(o,t) & Va.\& Ra.& OTE & Va.\& Ra.& Arhiliuc et al.& 0.447 & \\
        \cdashline{3-10}
         &  & Va.\& Ra. & Va.\& Ra. & wF1(o,t) &  Va.\& Ra. & OTE &  Va.\& Ra. & Bestgen & 0.647 & \\
        \cdashline{3-10}
         &  & Va.\& Ra. & Va.\& Ra. & wF1(o,t) &  Va.\& Ra. & OTE &  Va.\& Ra. & Bestgen & 0.696 & \\
        \cdashline{3-10}
         &  & Va.\& Ra. & Va.\& Ra. & wF1(o,t) &  $\approx$Va.\& Ra. & OTE &  Va.\& Ra. & Bestgen & 0.711 & \\
        \cdashline{3-10}
         &  & Va.\& Ra. & Va.\& Ra. & wF1(o,t) &  Va.\& Ra. & OTE &  Va.\& Ra. & Cai. \& But. & 0.647 & \\
        \cdashline{3-10}
         &  & Cai. \& But. & Cai. \& But. & wF1(o,t) & Cai. \& But. & OTE &  Va.\&Ra. & Cai. \& But. & 0.726 & \\
        \hline
        \multirow{8}{*}{mult-emb$^-$} & \multirow{8}{*}{wF1} & Va.\& Ra. & Va.\& Ra. & wF1(o,t) & Va.\& Ra.& OTE & Va.\& Ra.& Va.\& Ra.& 0.693 & \multicolumn{1}{c|}{\multirow{8}{*}{\textit{17.033}}}\\
        \cdashline{3-10}
         &  & Va.\& Ra. & Huber \& Coltekin& wF1(o,t) & Va.\& Ra.& OTE & Va.\& Ra.& Huber \& Coltekin& 0.658 & \\
        \cdashline{3-10}
         &  & Va.\& Ra.& Arhiliuc et al.& wF1(o,t) & Va.\& Ra.& OTE & Va.\& Ra.& Arhiliuc et al.& 0.683 & \\
        \cdashline{3-10}
         &  & Va.\& Ra. & Va.\& Ra. & wF1(o,t) &  Va.\& Ra. & OTE &  Va.\& Ra. & Bestgen & 0.668 & \\
        \cdashline{3-10}
         &  & Va.\& Ra. & Va.\& Ra. & wF1(o,t) &  Va.\& Ra. & OTE &  Va.\& Ra. & Bestgen & 0.692 & \\
        \cdashline{3-10}
         &  & Va.\& Ra. & Va.\& Ra. & wF1(o,t) &  $\approx$Va.\& Ra. & OTE &  Va.\& Ra. & Bestgen & 0.689 & \\
        \cdashline{3-10}
         &  & Va.\& Ra. & Va.\& Ra. & wF1(o,t) &  Va.\& Ra. & OTE &  Va.\& Ra. & Cai. \& But. & 0.659 & \\
        \cdashline{3-10}
         &  & Cai. \& But. & Cai. \& But. & wF1(o,t) & Cai. \& But. & OTE &  Va.\&Ra. & Cai. \& But. & 0.391 & \\
        \hline
        \multirow{8}{*}{mult-emb$^+$} & \multirow{8}{*}{wF1} & Va.\& Ra. & Va.\& Ra. & wF1(o,t) & Va.\& Ra.& OTE & Va.\& Ra.& Va.\& Ra.& 0.689 & \multicolumn{1}{c|}{\multirow{8}{*}{\textit{16.226}}}\\
        \cdashline{3-10}
         &  & Va.\& Ra. & Huber \& Coltekin& wF1(o,t) & Va.\& Ra.& OTE & Va.\& Ra.& Huber \& Coltekin& 0.662 & \\
        \cdashline{3-10}
         &  & Va.\& Ra.& Arhiliuc et al.& wF1(o,t) & Va.\& Ra.& OTE & Va.\& Ra.& Arhiliuc et al.& 0.681 & \\
        \cdashline{3-10}
         &  & Va.\& Ra. & Va.\& Ra. & wF1(o,t) &  Va.\& Ra. & OTE &  Va.\& Ra. & Bestgen & 0.659 & \\
        \cdashline{3-10}
         &  & Va.\& Ra. & Va.\& Ra. & wF1(o,t) &  Va.\& Ra. & OTE &  Va.\& Ra. & Bestgen & 0.681 & \\
        \cdashline{3-10}
         &  & Va.\& Ra. & Va.\& Ra. & wF1(o,t) &  $\approx$Va.\& Ra. & OTE &  Va.\& Ra. & Bestgen & 0.684 & \\
        \cdashline{3-10}
         &  & Va.\& Ra. & Va.\& Ra. & wF1(o,t) &  Va.\& Ra. & OTE &  Va.\& Ra. & Cai. \& But. & 0.657 & \\
        \cdashline{3-10}
         &  & Cai. \& But. & Cai. \& But. & wF1(o,t) & Cai. \& But. & OTE &  Va.\&Ra. & Cai. \& But. & 0.401 & \\
        \hline
\end{tabular}
\end{scriptsize}
    \caption{Conditions of measurement for each measurement carried out for the multilingual \textbf{essay scoring systems}. OTE = outputs vs.targets evaluation.}
    \label{tab:estc-conditions2} \label{tab:estc-conditions}
\end{table*}

\vspace{5cm}
\textcolor{white}{.}
\end{document}